\documentclass[letterpaper, 10 pt, conference]{ieeeconf}  % Comment this line out if you need a4paper
\usepackage{comment}
\usepackage{graphicx}
\usepackage{subcaption}
\usepackage{amsmath,amsfonts,amssymb, bm}
\usepackage{xcolor}
\definecolor{revised}{RGB}{0, 0, 0}
\usepackage[disable]{todonotes}
\makeatletter
\let\NAT@parse\undefined
\makeatother
\usepackage[colorlinks=true]{hyperref}

\IEEEoverridecommandlockouts                              % This command is only needed if 
\title{\LARGE \bf
NormalFlow: Fast, Robust, and Accurate Contact-based Object 6DoF Pose Tracking with Vision-based Tactile Sensors
}

\author{Hung-Jui Huang$^{1}$, Michael Kaess$^{1}$, and Wenzhen Yuan$^{2}$% <-this % stops a space
\thanks{$^{1}$Hung-Jui Huang and Michael Kaess are with Carnegie Mellon University, Pittsburgh, PA, USA 
        {\tt\small \{hungjuih, kaess\}@andrew.cmu.edu }}%
\thanks{$^{2}$Wenzhen Yuan is with University of Illinois Urbana-Champaign, Champaign, IL, USA
        {\tt\small yuanwz@illinois.edu }}%
}

\begin{document}

\maketitle
\thispagestyle{empty}
\pagestyle{empty}

%%%%%%%%%%%%%%%%%%%%%%%%%%%%%%%%%%%%%%%%%%%%%%%%%%%%%%%%%%%%%%%%%%%%%%%%%%%%%%%%
\begin{abstract}
Tactile sensing is crucial for robots aiming to achieve human-level dexterity. Among tactile-dependent skills, tactile-based object tracking serves as the cornerstone for many tasks, including manipulation, in-hand manipulation, and 3D reconstruction. In this work, we introduce NormalFlow, a fast, robust, and real-time tactile-based 6DoF tracking algorithm. Leveraging the precise surface normal estimation of vision-based tactile sensors, NormalFlow determines object movements by minimizing discrepancies between the tactile-derived surface normals. Our results show that NormalFlow consistently outperforms competitive baselines and can track low-texture objects like table surfaces. For long-horizon tracking, we demonstrate when rolling the sensor around a bead for $360$ degrees, NormalFlow maintains a rotational tracking error of $2.5$ degrees. Additionally, we present state-of-the-art tactile-based 3D reconstruction results, showcasing the high accuracy of NormalFlow. We believe NormalFlow unlocks new possibilities for high-precision perception and manipulation tasks that involve interacting with objects using hands. The video demo, code, and dataset are available on our website: \href{https://joehjhuang.github.io/normalflow}{https://joehjhuang.github.io/normalflow}.

\end{abstract}

\section{Introduction} \label{introduction}
\begin{comment}
tactile-based tracking is important
1.  Robots need to interact and manipulate with objects.
2. Knowing the object's pose is very important in manipulation.
3. While vision allows object tracking during interactions, interactions can occlude vision. For example, \cite{Openai2018} incorporates $19$ cameras to track in-hand object movements.
5. With the help of vision-based tactile sensors, we now have the capability to perform object tracking well under any occlusion and lighting conditions.

Despite its widespread use, vision-based object tracking often suffers from view occlusion during manipulation.
\end{comment}

\begin{figure}[t]
\centering
\includegraphics[width=0.98\linewidth]{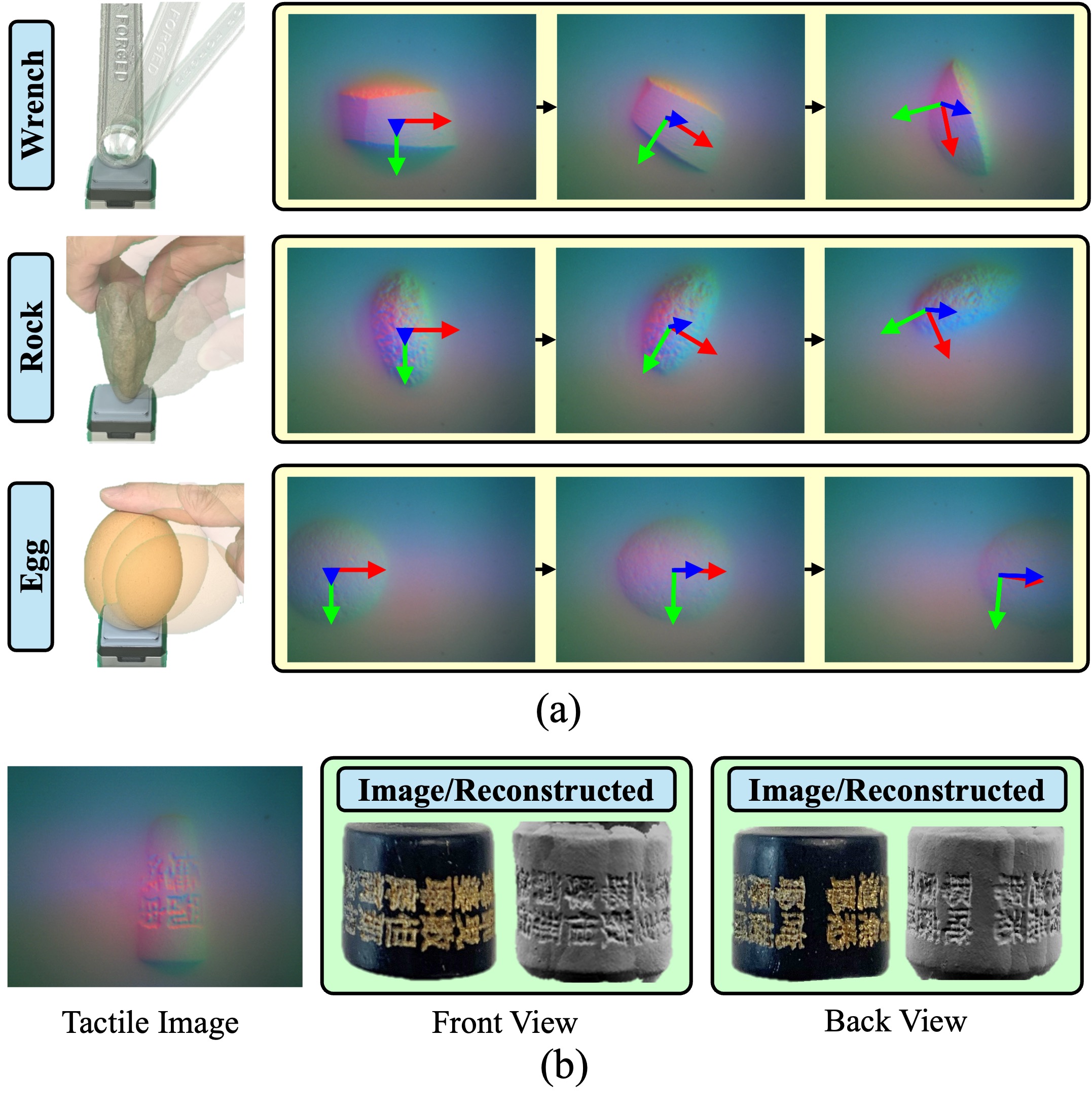}
\caption{NormalFlow performs fast, accurate, and robust 6DoF object tracking based on only touch sensing. (a) Accurate tracking of a wide variety of objects, including a wrench, a rock, and even low-texture object like an egg. (b) Applying NormalFlow to tactile-based 3D reconstruction of a $12$mm wide bead highlights NormalFlow's high accuracy.}
\label{fig:teaser}
\end{figure}

The skill to interact with and manipulate objects is fundamental for diverse robotics applications. For effective manipulation, accurate in-hand object tracking is critical, as it enables precise control based on the object's pose. Although vision-based systems are widely used for object tracking, they often suffer from occlusion during manipulation. OpenAI's dexterous manipulation system \cite{openai2018} highlights this challenge, as it uses $19$ cameras from every angle just to track a block's rotation in hand. Fortunately, the development of vision-based tactile sensors like GelSight \cite{yuan2017} offers a way to accurately track objects without occlusion issues.

In this work, we aim to accurately track objects during contact using vision-based tactile sensors without needing object 3D models. While this capability is the cornerstone of many downstream tasks such as manipulation, in-hand manipulation, and tactile-based 3D reconstruction, the field has not adequately addressed it. Works such as \cite{zhao2023} \cite{suresh2023} \cite{sodhi2022} \cite{Junyuan2023} focus on these downstream tasks and handle object tracking by converting tactile images into point clouds and applying registration methods like ICP \cite{Chen1991} to determine object transformations. However, these tracking methods often perform poorly with tactile-derived point clouds, which are often noisy and distorted. This issue arises because GelSight and its variants directly measure surface normals using photometric stereo methods \cite{johnson2011}, and converting these normals into point clouds requires integration, introducing accumulated noise and distortion. At its core, the field still lacks a fast, accurate, and robust tactile-based tracking solution.

In this paper, we present a novel tactile-based tracking algorithm, NormalFlow, which achieves significantly better robustness and accuracy than existing point cloud registration approaches and runs as fast or faster. Our key insight is to find object transformations by minimizing discrepancies between the surface normal maps of the contact area instead of point clouds. Since derived normal maps are much more accurate for vision-based tactile sensors, this approach yields more precise tracking results. For example, NormalFlow can accurately estimate object rolling angles by computing the rotations of normal vectors, which was previously challenging to estimate for point cloud registration approaches \cite{zhao2023}. Inspired by the Lucas-Kanade optical flow method \cite{Lucas1981}, NormalFlow uses Gauss-Newton optimization to minimize discrepancies between normal maps and determine the 6DoF object pose quickly, robustly, and accurately. Additionally, we propose a robust long-horizon tracking method based on NormalFlow to reduce drifts under extended movements.

We test NormalFlow's tracking performance on a wide variety of objects and show it consistently outperforms point cloud registration baselines. {\color{revised} It achieves an average translation error of $0.29$mm (over a total movement of $3.4$mm), rotation error of $1.9^\circ$ (over a total movement of $37.4^\circ$), and runs at $70$Hz on CPU.} NormalFlow can even track objects with low textures, like a flat table surface. For long-horizon tracking, NormalFlow maintains a rotational tracking error of $2.5^\circ$ after rolling the sensor around a $12$mm wide bead for $360^\circ$. Additionally, we show NormalFlow's generalizability across different sensors and resolutions. Finally, to demonstrate NormalFlow's power, we apply it to tactile-based 3D reconstruction. Leveraging NormalFlow's high precision, our reconstructed geometry significantly outperforms the geometry reconstructed using motion capture system tracked poses. We believe NormalFlow opens new avenues for higher precision tactile-dependent perception and control.

\section{Related Work} \label{related_work}
% Tactile sensing is important for robots aiming to achieve human-level dexterity, with one crucial aspect being improving object pose estimation during interaction. 
Early methods for tactile-based object pose estimation focused on low-resolution tactile sensors \cite{anna2011}, often requiring known 3D object models and numerous tactile readings. Alternatively, most recent approaches \cite{sodhi2022} \cite{suresh2022} \cite{bauza2023} utilize high-resolution sensors such as GelSight and its variants \cite{johnson2011} \cite{yuan2017} \cite{lambeta2020}, achieving more efficient and accurate pose estimation. 
This review focuses on these recent methods and their achievements in pose estimation tasks.
%This review will focus on these recent advancements and discuss their achievements in the three main categories of pose estimation tasks.

%Early methods for tactile-based object pose estimation focused on low-resolution tactile sensors \cite{anna2011} \cite{javdani2013}. These approaches often required known 3D object models and many tactile readings. Alternatively, most recent works \cite{suresh2022} \cite{sodhi2022} \cite{bauza2023} rely on high-resolution tactile sensors such as vision-based tactile sensors to achieve a more efficient and accurate pose estimation results. In the following, we focus on reviewing those works.

%\noindent \textbf{Vision-based Tactile Sensors:} Vision-based tactile sensors \cite{johnson2011} \cite{yuan2017} \cite{lambeta2020} are a class of robust, low-cost, high-resolution tactile sensors. They typically include a soft sensing unit that deforms when in touch, an illumination unit that 

\noindent \textbf{Frame-to-Frame Tracking:} Frame-to-frame tracking is to track an object's movement in the sensor frame using only tactile data. It is the prerequisite for many tactile-dependent tasks \cite{Rui2014} \cite{zhao2023} \cite{suresh2023}. To determine objects' planar movements in $SE(2)$, Li et al. \cite{Rui2014} used a feature-based image registration method on tactile-derived height maps, while Sodhi et al. \cite{sodhi2020} trained a neural network on tactile images. A more important task is determining 6DoF object movement in $SE(3)$. Researchers often reduce this problem to registering tactile-derived point clouds or height maps. For instance, \cite{suresh2023} \cite{sodhi2022} \cite{wang2021} used ICP to register point clouds and determine object movements, while \cite{Junyuan2023} performed feature-based point cloud registration before applying ICP. Zhao et al. \cite{zhao2023} trained a neural network to predict object transformations from tactile-based height maps. However, these approaches are limited by the often distorted point cloud quality, leading to noisy tracking results and sometimes making rolling movements impossible to estimate \cite{zhao2023}. The working principle of vision-based tactile sensors is to use photometric stereo methods to acquire geometrical information \cite{johnson2011}, specifically through the estimation of surface normals. Deriving point clouds, in this case, by integration of surface normals will result in accumulated noise. Our approach directly measures object movements using the surface normals, leading to more accurate estimations. %For example, objects' rolling movement can be determined by the rotation of normal vectors.

\noindent \textbf{Long-horizon Tracking:} Long-horizon tracking is usually achieved by composing frame-to-frame tracking results, which can introduce drift. To address this, most methods applied factor graph optimization \cite{zhao2023} \cite{kim2024}. {\color{revised} For example, Sodhi et al. \cite{sodhi2022} built a factor graph upon the ICP frame-to-frame tracking method, reporting $4$ mm translation and $0.2$ rad rotation errors when tracking a pyramid-shaped object.} %In comparison, our method achieves a $1.5$ mm translation and a $0.04$ rad rotation error over a full $6.28$ rad rolling on a bead.
%Frame-to-frame tracking allows us to determine long-horizon object movements while in contact. To address drift from composing frame-to-frame tracking results,

%\noindent \textbf{Global Localization:} The goal of this task is to determine the touch location given the object 3D model in prior. Suresh et al. \cite{suresh2022} achieved this on YCB objects \cite{Calli2017} using particle filters with measurement models on learned representation. Bauza et al. \cite{bauza2023} estimated the distribution of touch locations on small objects based on learned tactile representation. Another common approach is to combine vision and touch, but this is beyond the scope of our discussion.
\section{Method} \label{method}
%We aim to track the 6DoF object poses using vision-based tactile sensors without requiring the objects' geometrical models. Utilizing photometric stereo methods, surface normal maps are derived from the stream of tactile images \cite{wang2021}. The key insight of our NormalFlow algorithm is to find the transformation by directly minimizing discrepancies between normal maps rather than point clouds. This method is less susceptible to noise compared to using point clouds, which are noisier due to the integration process needed to generate them from normal maps. In contrast, existing approaches \cite{sodhi2022} \cite{bauza2023} that use ICP or FilterReg \cite{gao2019} to register point clouds are more prone to this noise accumulation.
% We aim to track the 6DoF object poses using vision-based tactile sensors without requiring the objects' geometrical models. 
We aim to use vision-based tactile sensors to track the 6DoF object poses relative to the sensor's coordinate frame during contact, without requiring the objects' geometrical models. Utilizing photometric stereo methods, surface normal maps are derived from the stream of tactile images \cite{wang2021}. The key insight of our NormalFlow algorithm is to find the transformation by directly minimizing discrepancies between normal maps rather than point clouds. This method prevents the accumulated noise associated with the integration process when converting normal maps to point clouds. In contrast, existing approaches \cite{sodhi2022} \cite{bauza2023} used ICP or FilterReg \cite{gao2019} to register point clouds and are prone to noise accumulation.

\subsection{Surface Geometries Prediction from Tactile Images}
Vision-based tactile sensors like GelSight \cite{johnson2011} \cite{yuan2017} and its derivatives \cite{lambeta2020} are designed with hardware that employs photometric stereo methods \cite{johnson2011} to capture surface normals at each individual pixel from the sensor's reading. We adapt the approach from \cite{wang2021} to derive surface normal maps from tactile images. {\color{revised} The method is chosen for its simplicity, widespread use \cite{zhao2023} \cite{Junyuan2023}, and accuracy, and can be replaced with alternative approaches \cite{bauza2023} \cite{suresh2022}. Using the approach in \cite{wang2021}, we collect $50$ tactile images by pressing a $5$mm diameter metal ball against various regions of the sensor's sensing surface. By manually labeling the circular contact region, we calculate the true surface gradients of each pixel within these regions based on the known ball diameter.} A multi-layer perceptron with three hidden layers (5-128-32-32-2) is then trained to learn a mapping from the pixel's color and position (RGBUV) to its surface gradient ($g_u$, $g_v$). During test time, the surface normal $\hat{\mathbf{n}}$ of each pixel is computed from the predicted surface gradient by $\hat{\mathbf{n}} = \mathbf{n}/ \|\mathbf{n}\|$, where $\mathbf{n} = \begin{bmatrix}g_u & g_v & -1\end{bmatrix}^\intercal$.

To derive the height map $z$, a 2D fast Poisson solver \cite{yuan2017} is applied to integrate the predicted surface gradients. This process accumulates noise and distortion and can hardly be mitigated due to the design choice of the sensor hardware, which directly perceives normals instead of heights. The contact region is straightforwardly determined by regions where the height and tactile image variations exceed a threshold compared to pre-contact conditions.
%While the primary focus of the NormalFlow algorithm is on matching surface normal maps, intermediate steps still require computing the height map and contact mask.
%This gradient-to-height integration accumulates noise and distortion, which explains why baseline methods that purely rely on point cloud registration are less stable and less accurate.

\begin{figure}[t]
\centering
\includegraphics[width=0.98\linewidth]{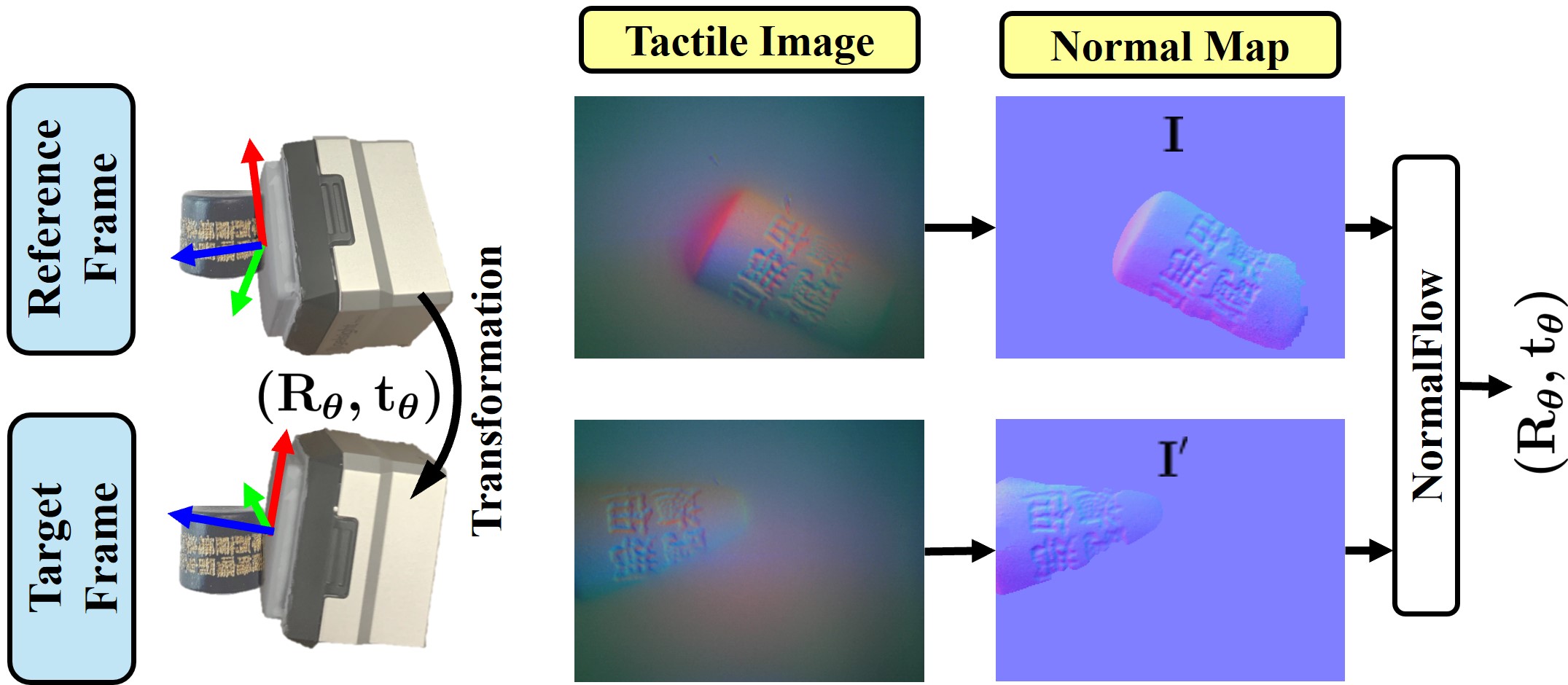}
\caption{Given two tactile images before and after object movement, we derive the surface normal maps. NormalFlow determines the object transformations by minimizing discrepancies between the surface normal maps.}
%\caption{NormalFlow finds the transformation that minimizes the normal discrepancies. The normal discrepancy at pixel $(x,y)$ is computed as the difference between the rotated reference normal $\mathbf{R}_{\bm{\theta}}\mathbf{I}(x,y)$ at that pixel and the target normal $\mathbf{I}'(\mathbf{W}(x,y;\bm{\theta}))$ at the warped pixel coordinate. Warping is done by transforming the pixel's corresponding 3D surface coordinate $\mathbf{q}(x,y)$ and projecting it back to the x-y plane.}\joe{Maybe magnify the gradient so normals are more obvious. Meanwhile, manually shift or rotate the target surface normal map so the warping is more obvious.}\wenzhen{I think it's better to have a photo of external view to get people quickly understand the problem setting. Also, the side view of the normal form is hard to see. It's better to used the front view, and better to use the raw image. I think you can draw the contact frame on the 3D object's model. Also, for figure notation, it's a bad idea to use white text. Ye}
\label{fig:method_figure}
\end{figure}

% Consider a reference sensor frame and a target sensor frame, each characterized by their respective perceived surface normal maps $\mathbf{I} \text{ and } \mathbf{I}':\mathbb{R}^2 \mapsto \mathbb{R}^3$, mapping from pixel coordinates to surface normals.

\subsection{The NormalFlow Algorithm}
Consider a reference sensor frame and a target sensor frame, each characterized by their respective perceived surface normal maps $\mathbf{I} \text{ and } \mathbf{I}'$ as shown in Fig. \ref{fig:method_figure}, which are functions $\mathbb{R}^2 \mapsto \mathbb{R}^3$ mapping pixel coordinates to surface normals. Our goal is to estimate the 6DoF transformation from the reference frame to the target frame $(\mathbf{R}_{\bm{\theta}}, \mathbf{t}_{\bm{\theta}})\in SE(3)$, parameterized as $\bm{\theta}=(x, y, z, \theta_x, \theta_y, \theta_z) \in \mathbb{R}^6$. The NormalFlow algorithm minimizes the difference between two surface maps within the shared contact region: the transformed reference map $\mathbf{I}$ when applying the 3D transformation to the object, and the target map $\mathbf{I}'$. This 3D transformation affects the reference map $\mathbf{I}$ by rotating the normal directions and re-mapping the pixels. Therefore, the minimization objective of NormalFlow is:

\begin{equation} \label{eq:loss}
\sum_{(u,v)\in \overline{C}} \; [\mathbf{I}'(\mathbf{W}(u,v;\bm{\theta})) - \mathbf{R}_{\bm{\theta}}\mathbf{I}(u,v)]^2
\end{equation}
where $(u,v)$ is the pixel coordinates. The shared contact region $\overline{C}$ is computed as the intersection of the target frame's contact region and the re-mapped contact region in the reference frame. The re-mapping function $\mathbf{W}(u, v; \bm{\theta})$ maps pixel coordinates from the reference frame to the target frame by transforming the 3D surface coordinate of the pixels and projects it to 2D:
%where $(u,v)$ is the pixel coordinates and $\overline{C}$ denotes the shared contact region between the two normal maps.

%The NormalFlow algorithm minimizes the squared difference between two surface normal maps within the shared contact region: the reference map $\mathbf{I}$ with rotated normals and the target map $\mathbf{I}'$ warped back onto the reference map's coordinate frame:

\begin{equation} \label{eq:warp}
\mathbf{W}(u,v;\bm{\theta}) = \mathbf{P}\big(\mathbf{R}_{\bm{\theta}}\cdot\mathbf{q}(u,v) + \mathbf{t}_{\bm{\theta}}\big)
\end{equation}
where $\mathbf{q}(u,v) = \begin{bmatrix}u & v & z(u,v)\end{bmatrix}^\intercal$ is the 3D surface coordinate corresponding to the pixel at $(u, v)$ in the reference frame and $\mathbf{P} = (\begin{smallmatrix} 1 & 0 & 0 \\ 0 & 1 & 0 \end{smallmatrix})$ is the projection matrix. Since the indentation is always small ($z(u,v) \ll u, v$), the noise of the height estimation $z(u,v)$ has minimal impact on the re-mapping function. Inspired by the Lucas-Kanade optical flow method \cite{Lucas1981} \cite{Baker2004}, NormalFlow employs the Gauss-Newton optimization to minimize Eq. (\ref{eq:loss}) iteratively. For clarity, we simplify notation by omitting arguments. Linearizing Eq. (\ref{eq:loss}) at the current estimate of $\bm{\theta}$ results in:

\begin{equation} \label{eq:linearized_loss}
\sum_{(u,v)\in \overline{C}}\big[
\big(\mathbf{I}'(\mathbf{W}) - \mathbf{R}_{\bm{\theta}}\mathbf{I}\big) + 
\big(\nabla \mathbf{I}' \frac{\partial \mathbf{W}}{\partial \bm{\theta}} - \frac{\partial(\mathbf{R}_{\bm{\theta}}\mathbf{I})}{\partial \bm{\theta}}\big)\bm{\mathit{\Delta}} \bm{\theta}\big]^2
\end{equation}
where $\nabla \mathbf{I}' = [\frac{\partial \mathbf{I}'}{\partial u}|\frac{\partial \mathbf{I}'}{\partial v}]$ is the gradient normal map evaluated at the re-mapped pixel coordinate $\mathbf{W}(u,v;\bm{\theta})$. 
%The closed-form expression of the Jacobian of $\mathbf{W}$ and $\mathbf{R}_{\bm{\theta}}\mathbf{I}$ are presented in Appendix \ref{appendix:jacobian}.
We can further write Eq. (\ref{eq:linearized_loss}) in the form of $\left\lVert\mathbf{A}\bm{\mathit{\Delta}}\bm{\theta} - \mathbf{b}\right\rVert^2$, representing a linear least square problem. The closed-form solution of it is $\bm{\mathit{\Delta}} \bm{\theta} = \mathbf{H}^{-1}\mathbf{A}^\intercal \mathbf{b}$, where $\mathbf{H} = \mathbf{A}^\intercal \mathbf{A}$ is the Hessian matrix. We update the parameters by $\bm{\theta} \gets \bm{\theta} + \bm{\mathit{\Delta}} \bm{\theta}$ and repeat the linearization and parameters update process until the parameters converge.

\begin{comment}
\begin{equation}\label{eq:forward_update}
\bm{\theta} \gets \bm{\theta} + \bm{\mathit{\Delta}} \bm{\theta}
\end{equation}
\end{comment}

Minimizing discrepancies of the surface normal maps can not resolve the z-translation component of $\bm{\theta}$ due to their 2D nature. This limitation stems from the 3D to 2D projection involved in the pixel re-mapping process, which results in the Hessian matrix with zero values for the z-translation dimension. Therefore, we calculate the z-translation independently once the other five dimensions are determined using Gauss-Newton optimization. We do this by computing, within the shared contact region, the mean difference between the target height map and the reference height map transformed using the parameters $\bm{\theta}$ determined by the Gauss-Newton optimization:
\begin{equation}
\Delta z = \frac{1}{|\overline{C}|}\sum_{(u,v)\in\overline{C}} [z'(\mathbf{W}) - \mathbf{P}_z(\mathbf{R}_{\bm{\theta}}\cdot\mathbf{q} + \mathbf{t}_{\bm{\theta}})]
\end{equation}
Here, $z'(\mathbf{W})$ represents the target height map at the re-mapped pixel coordinate, and $\mathbf{P}_z = (\begin{smallmatrix}0 & 0 & 1\end{smallmatrix})$ represents the z-axis projection matrix. We then replace the z-translation dimension in $\bm{\theta}$ with the calculated $\Delta z$ to finalize the transformation estimate. Despite relying on the noisy height map $z$ and $z'$, $\Delta z$ estimation errors remain small because $\Delta z$ itself is always small for the object to maintain contact.

\subsection{Inverse Composition and Random Subsampling} \label{section:speed_up}
We employ two techniques to speed up NormalFlow without compromising its performance. During each iteration of the Gauss-Newton optimization, the Hessian matrix $\mathbf{H}$ is re-evaluated at an updated parameter $\bm{\theta}$. We apply the inverse compositional method \cite{Baker2004} to reformulate Eq. (\ref{eq:linearized_loss}) so that the Hessian matrix can be pre-computed and re-used for every iteration (see Appendix \ref{appendix:inverse_compositional}). This approach lowers the computational complexity from $O(nm^2N)$ to $O(m^2N+nmN)$, where $n$ is the number of iterations, $m=6$ is the parameter dimension, and $N$ is the number of pixels in the shared contact region. Additionally, to further reduce runtime, we randomly subsample $N=5000$ pixels in the shared contact region from a tactile image resolution of $320 \times 240$ when evaluating Eq. (\ref{eq:linearized_loss}). {\color{revised} This choice of $N$ serves as a conservative lower bound that ensures negligible impact on result quality in practice. Smaller values of $N$ sometimes still work but offer diminishing returns on runtime reduction.}
%, which affects result quality negligibly. {\color{revised} Smaller choices of $N$ can still work but offer diminishing returns on runtime reduction.}

%Additionally, we random subsample pixels within the shared contact region when evaluating Equation (\ref{linearized_loss}) to reduce the number $N$. In practice, we subsample $N=5000$ pixels when the tactile image has $320\times 240$ resolution since that has negligible effect on the quality of the result.

\subsection{Long-Horizon Tracking}
We propose a keyframe-based approach for long-horizon tracking. Unlike naive approaches that compute transformations relative to the immediately preceding frame and compose these transformations over the horizon, our method computes the transformation from the current frame to the latest keyframe and composes it with the aggregated transformations between keyframes. This technique minimizes drift by reducing the number of transformation composition needed, leveraging the infrequent occurrence of keyframes compared to regular frames. During tracking, if a frame is far from the latest keyframe causing large NormalFlow estimation errors, it is set as the new keyframe. Specifically, using normal maps $\mathbf{I}_k$, $\mathbf{I}_p$, and $\mathbf{I}_c$ from the latest keyframe, previous, and current frames, we estimate transformations via NormalFlow. We calculate $^c\mathbf{T}_k$ between the latest keyframe and current frame, and $^c\mathbf{T}_p$ between the previous and current frames. If the difference between $({}^c\mathbf{T}_p)^{-1} \cdot {}^c\mathbf{T}_k$ and $^p\mathbf{T}_k$ (estimated in the previous time step) exceeds a threshold, we set the previous frame as the new keyframe. {\color{revised} To further reduce drift, factor graph optimization \cite{sodhi2022} can be built upon our approach; however, this is beyond the scope of our paper.}

%We propose a keyframe-based approach to achieve long-horizon tracking. Instead of computing the transformation relative to the frame in the previous time step and integrating over the horizon, we propose to compute the transformation relative to the most recent keyframe and integrates with the keyframe-to-keyframe transformation. Since there are much fewer keyframes than frames, the keyframe-based approach drifts slower by perform less integration.

%The NormalFlow algorithm determines the transformation between two frames. For long-horizon tracking, we propose to keep the reference frame the same and compute transformation relative to the same reference frame unless transformation is already too large to track and a new key frame is needed. Compared to the naive way of always setting the reference frame as the frame collected at the previous time step and integrates the transformation to compute the transformation with regards to the initial frame, this method drifts slower. Specifically, we determine a new keyframe is needed by running NormalFlow twice. Consider the current reference frame being

\subsection{Discussion and Comparison}

For math completeness, our NormalFlow derivation uses the noisy height map. In fact, the noise's effect on NormalFlow is minimal. In practice, {\color{revised} skipping height map computation by simply assuming a zero height map ($z(u,v) = 0$) has a minor effect on NormalFlow's estimation error} and still significantly outperforms the point cloud registration methods (See Section \ref{section:tracking_result} {\color{revised} for ablation results}).

%For math completeness, our derivation of NormalFlow uses the noisy height map, but the effect of its noise is actually minimal. skipping height map computation by naively setting $z(u,v)=0$ minorly affects the NormalFlow's estimation error and still significantly outperforms the point cloud registration baselines (See Section \ref{section:tracking_result}).

%For math completeness, our derivation of NormalFlow uses the noisy height map, but its effect is actually minimal. In practice, we find that skipping height map computation by naively setting $z(u,v)=0$ affects the NormalFlow's estimation error (Section \ref{section:tracking_result}) by only $9\%$.

Next, we discuss the advantages of using surface normals (NormalFlow) for pose estimation over point clouds (ICP). Consider determining the tilt of a flat surface. The rotations of surface normals directly reveal the tilt, whereas estimating tilt using distorted point clouds can be noisy. For a textured ball, pose estimation should rely on the textures. Unfortunately, ICP will focus on registering the global ball shape since the textures minimally affect point locations. In contrast, NormalFlow will estimate pose from textures by matching the diverse normal directions in the textured region.

\begin{comment}
The major advantage of NormalFlow is its avoidance of using the noisy height map estimates. While 
NormalFlow is much more precise than the point cloud registration approaches. Objects can be considered as containing global geometries and local features. Consider the most simple global geometry, a flat surface. When tilted, NormalFlow can estimate the tilting angle by estimating the rotation of the normals while point cloud registration approaches will integrate the normals into a distorted surface where the rotation estimation can be noisy. 
While the noisy height map estimations are involved in NormalFlow in the pixel re-mapping and $\Delta z$ estimation steps, it minimally impacts these process as discussed earlier. In fact, we show that naively using $z(u,v)=0$ has nearly no effect on the estimation error (Section 5); while point cloud registration methods will definitely not work for that.
The main advantage of our approach is it avoids the noisy height map estimation $z$. 
For mathematical completion, we derive NormalFlow with the noisy height map $z$.  
\end{comment}
\section{Experiments and Results} \label{experiments}
In this section, we perform experiments to evaluate the accuracy and speed of the NormalFlow method for tracking the motion of different objects. We also evaluate the effect of object movement speed on tracking performance and the method's generalizability with different tactile sensors and sensor resolutions. Unless specified, all experiments are conducted on the GelSight Mini sensor with a resized resolution of $320 \times 240$. This sensor has a $20 \text{mm} \times 15 \text{mm}$ sensing area and operates at 25 Hz.

% In the experiment section, we want to answer the following questions: (1) How does NormalFlow's accuracy compare to baseline methods when tracking different objects? (2) How does NormalFlow's processing speed compare to baselines? (3) What is the maximum object moving speed NormalFlow can track? (4) How does tactile resolution affect NormalFlow's accuracy? (5) Can NormalFlow generalize to other common vision-based tactile sensors? (6) How does NormalFlow perform in long-horizon tracking? Unless specified, all experiments are conducted on the GelSight Mini sensor with a resized resolution of $320 \times 240$. This sensor has a $20 \text{mm} \times 15 \text{mm}$ sensing area and operates at 25 Hz.

\subsection{Object 6DoF Pose Tracking}
In this experiment, we evaluate the tracking accuracy and runtime of NormalFlow against baseline methods on various objects and show its long-horizon tracking performance.
% \wenzhen{It might be better to make the instruction like subsection 1: experiment setup -- talking about object dataset and how you perform the experiments; subsection 2: baseline methods; subsection 3: 6D pose tracking -- where you talk about accuracy and runtime}
\subsubsection{Data Collection}\label{section:data_collection}
\begin{figure}[t]
\centering
\includegraphics[width=0.7\linewidth]{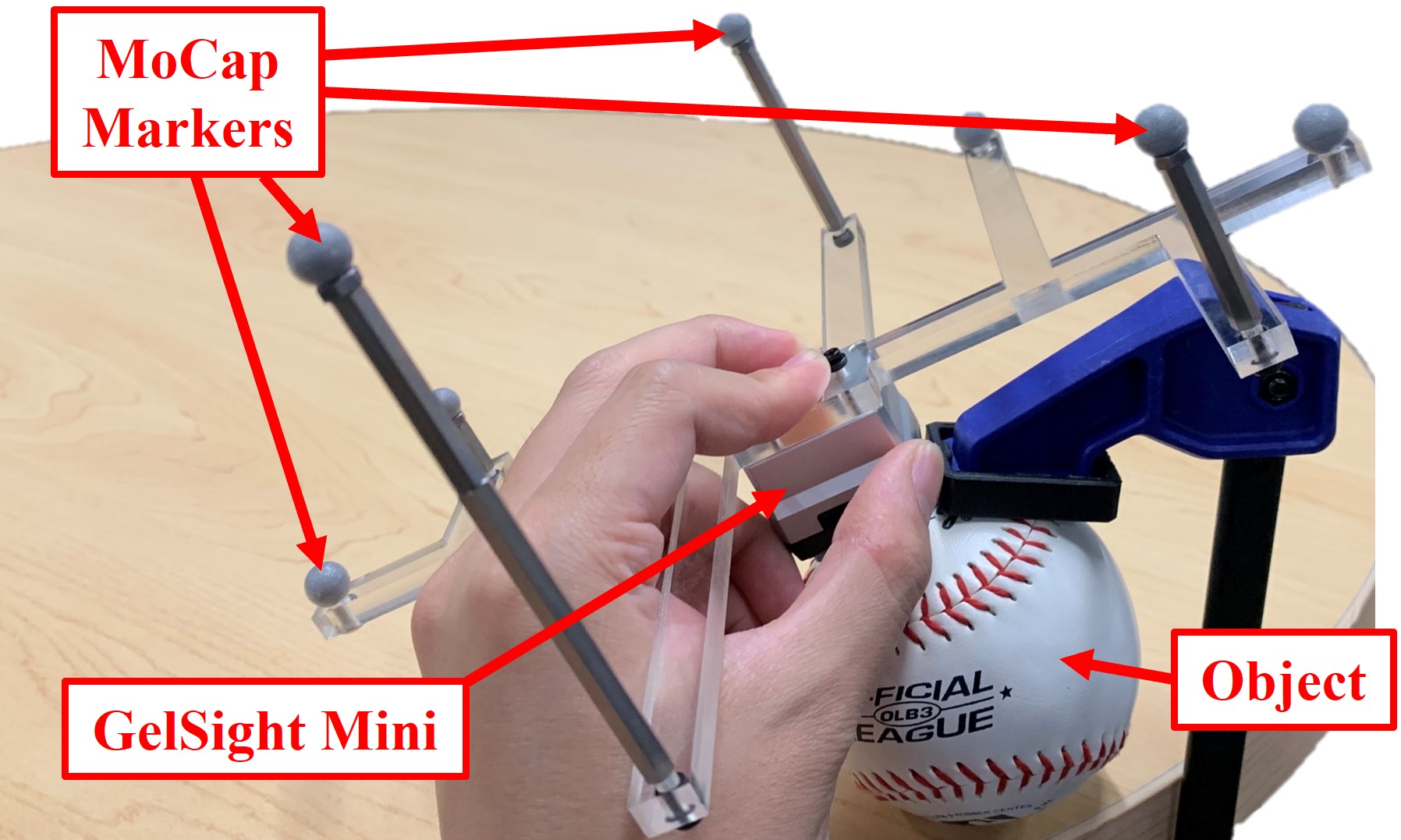}
\caption{The data collection setup, with the object clamped on the table and the GelSight sensor tracked by MoCap.}

\label{fig:setup}
\end{figure}

\begin{figure}[t]
\centering
\includegraphics[width=0.95\linewidth]{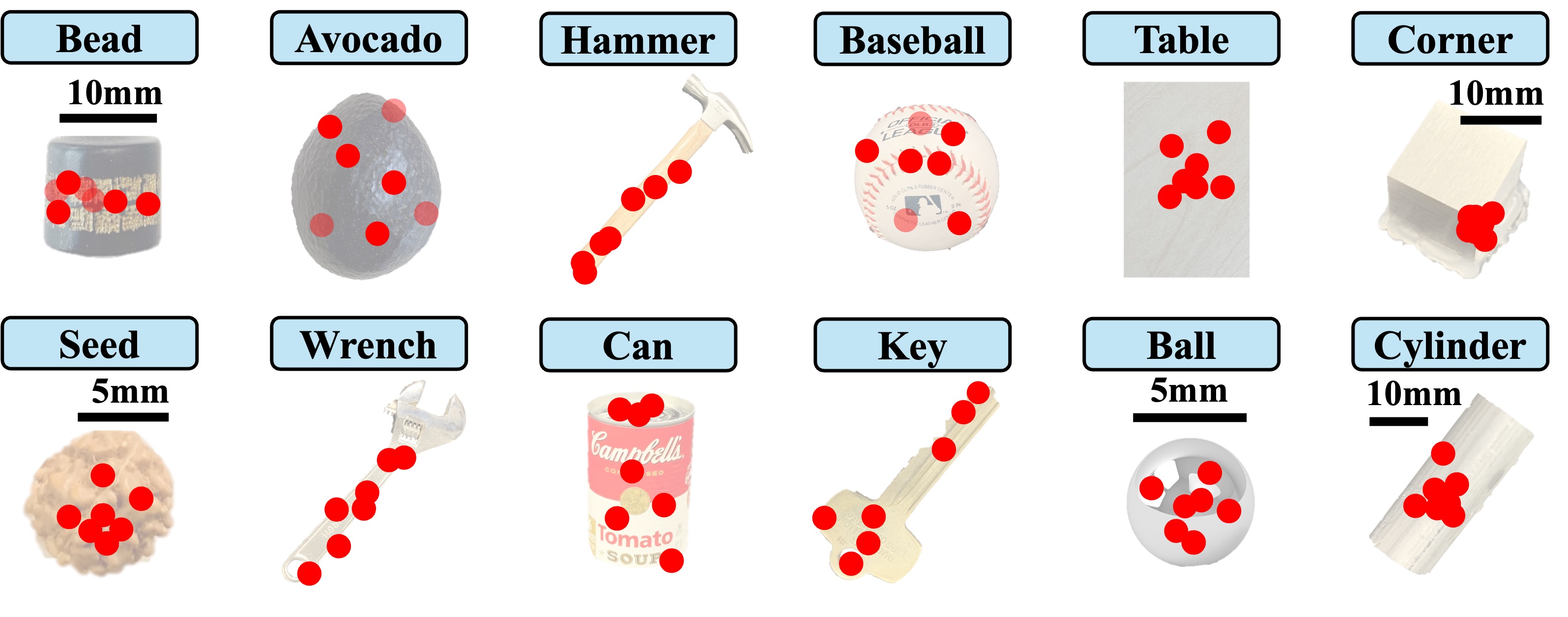}
\caption{Initial contact locations (manually labeled) for the seven trials per object in the tracking experiment.}
\label{fig:contact_locations}
\end{figure}

\begin{figure*}[htbp]
\centering
\includegraphics[width=0.98\linewidth]{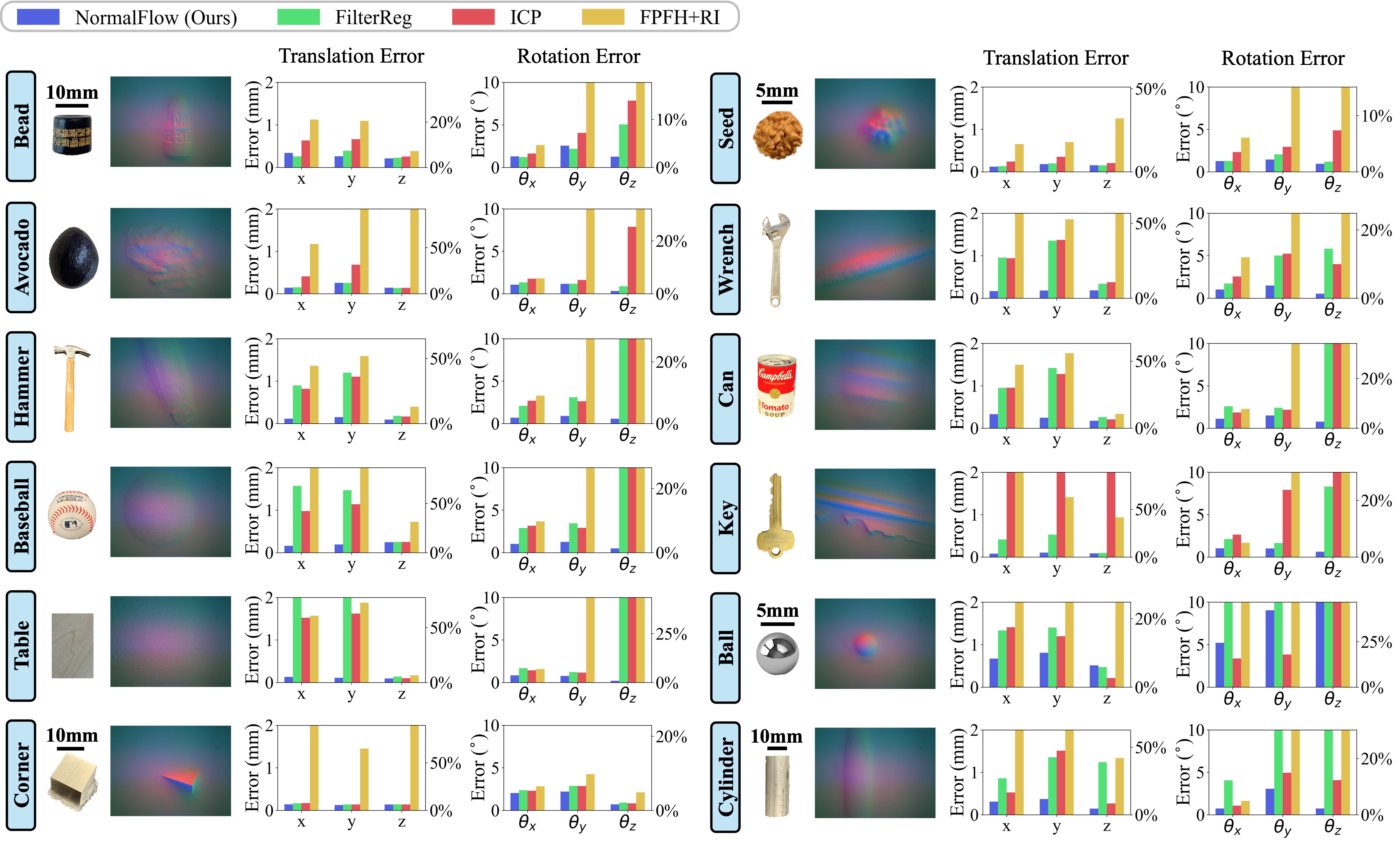}
\caption{Tracking results for the 12 objects. For each object: \textbf{[left]} the object (scale not shown for common objects) and a sample tactile image (Seed's image slightly blurred to avoid visualization discomfort); \textbf{[right]} {\color{revised} the 6DoF tracking MAE: left y-axis shows absolute error, right y-axis shows percentage error relative to object movement range.}}
\label{fig:tracking_result}
\end{figure*}

To evaluate our approach, we collect pose tracking data on $12$ objects in three categories: seven everyday objects (four from the YCB dataset \cite{Calli2017}), two small textured objects, and three simple geometric shapes (Fig. \ref{fig:tracking_result}). We clamp the objects on a fixed table and mount the GelSight Mini on a movable plate tracked by the OptiTrack motion capture system (MoCap) to capture the ground truth sensor poses (Fig. \ref{fig:setup}). {\color{revised}For each object, we collect seven tracking data trials, initiating contact at different poses shown in Fig. \ref{fig:contact_locations}.} To prevent quick damage to the Gel, which is very vulnerable to wearing under pure sliding actions, we mainly apply rotational movements on the object's surface, including rolling in the x-y plane and twisting along the z-axis. By changing the rotation center, rotational movements can cause object translations. Our data collection process includes these translations to demonstrate our approach's effectiveness in tracking object movements in all 6DoF. For each trial, we maintain the object's position not too far from its initial pose to ensure sufficient overlap in the contact area relative to the initial frame, thereby avoiding setting new keyframes. On average, each trial includes $10.2$sec of tracking data, and the average 6DoF moving range is shown in Table \ref{tab:moving_range}.
%On average, each trial includes $87$ tactile images, and the average 6DoF moving range is shown in Table \ref{tab:moving_range}.

%\textbf{[left]} The $12$ objects for evaluation. Detailed object shapes are shown in Fig. \ref{fig:tracking_result}. \textbf{[right]} The data collection setup.

%the object and a sample tactile image showing its texture (Seed's tactile image is slightly blurred to prevent potential visualization discomfort);

\begin{comment}
\begin{table}[ht]
\centering
\begin{tabular}{|c|c|c|c|c|c|c|}  %<-- change here
\hline 
Object & x(mm) & y(mm) & z(mm) & $\theta_x(^{\circ})$ & $\theta_y(^{\circ})$ & $\theta_z(^{\circ})$ \\ 
\hline 
\hline 
Seed & 1.74 & 2.85 & 1.89 & 32.9 & 32.2 & 47.7 \\ 
\hline 
Bead & 2.71 & 4.28 & 1.7 & 18.4 & 27.9 & 45.5 \\ 
\hline 
Avocado & 1.74 & 1.07 & 0.92 & 11.8 & 13.7 & 25.6 \\ 
\hline 
Wrench & 2.32 & 2.14 & 1.33 & 16.3 & 22.9 & 27.7 \\ 
\hline 
Hammer & 2.27 & 1.97 & 0.89 & 16.8 & 17.7 & 27.6 \\ 
\hline 
Can & 2.41 & 1.79 & 1.03 & 13.3 & 13.8 & 22.0 \\ 
\hline 
Baseball & 1.61 & 1.42 & 0.85 & 16.4 & 14.7 & 30.7 \\ 
\hline 
Key & 1.44 & 1.54 & 0.79 & 15.8 & 13.3 & 26.1 \\ 
\hline 
Table & 1.57 & 2.01 & 0.55 & 7.34 & 7.46 & 20.5 \\ 
\hline 
Ball & 6.04 & 5.38 & 0.89 & 12.1 & 12.5 & 12.0 \\ 
\hline 
Corner & 1.21 & 1.54 & 1.38 & 21.1 & 20.8 & 32.1 \\ 
\hline 
Cylinder & 2.53 & 1.66 & 1.05 & 12.0 & 19.5 & 24.2 \\ 
\hline 
\end{tabular}
\caption{Dataset statistics: moving range for each object.}
\label{tab:moving_range}
\end{table}
\end{comment}

\begin{table}[ht]
\centering
\begin{tabular}{|c|c|c|c|c|c|}  %<-- change here
\hline 
x(mm) & y(mm) & z(mm) & $\theta_x(^{\circ})$ & $\theta_y(^{\circ})$ & $\theta_z(^{\circ})$ \\ 
\hline 
\hline 
2.30 & 2.30 & 1.11 & 16.2 & 18.0 & 28.5 \\ 
\hline
\end{tabular}
\caption{Dataset statistics: average 6DoF moving range.}
\label{tab:moving_range}
\end{table}

\subsubsection{Baseline Methods}
We compare NormalFlow with three common point cloud registration methods. For each tactile image, we convert the predicted height map $z$ into a point cloud, excluding points outside the predicted contact region, which then serves as input for these baseline methods:

\noindent \textbf{Point-to-Plane ICP} \cite{Chen1991}: Referred to as ICP in this paper, Point-to-plane ICP is a local registration approach and is commonly used for tactile-based tracking \cite{wang2021} \cite{sodhi2022} \cite{suresh2023}. We utilize the implementation from the Open3D Library \cite{Zhou2018}.

\noindent \textbf{FilterReg} \cite{gao2019}: FilterReg is a probabilistic local point cloud registration approach. It is more robust than ICP and has been applied in \cite{Bauza2020} for tactile registration. We utilize the implementation from the ProbReg Library \cite{probreg2019}.

\noindent \textbf{FPFH + RANSAC + ICP}: Abbreviated as FPFH+RI in this paper, it combines Fast Point Feature Histograms (FPFH) \cite{Rasu2009} and RANSAC to extract features and register point clouds globally. It then applies ICP to refine the alignment. The approach is applied in \cite{Junyuan2023} for tactile registration. We utilize the implementation from the Open3D Library \cite{Zhou2018}.

The local registration methods (NormalFlow, ICP, and FilterReg) are initialized using the pose estimated from the previous frame.

\begin{figure}[t]
\centering
\includegraphics[width=0.98\linewidth]{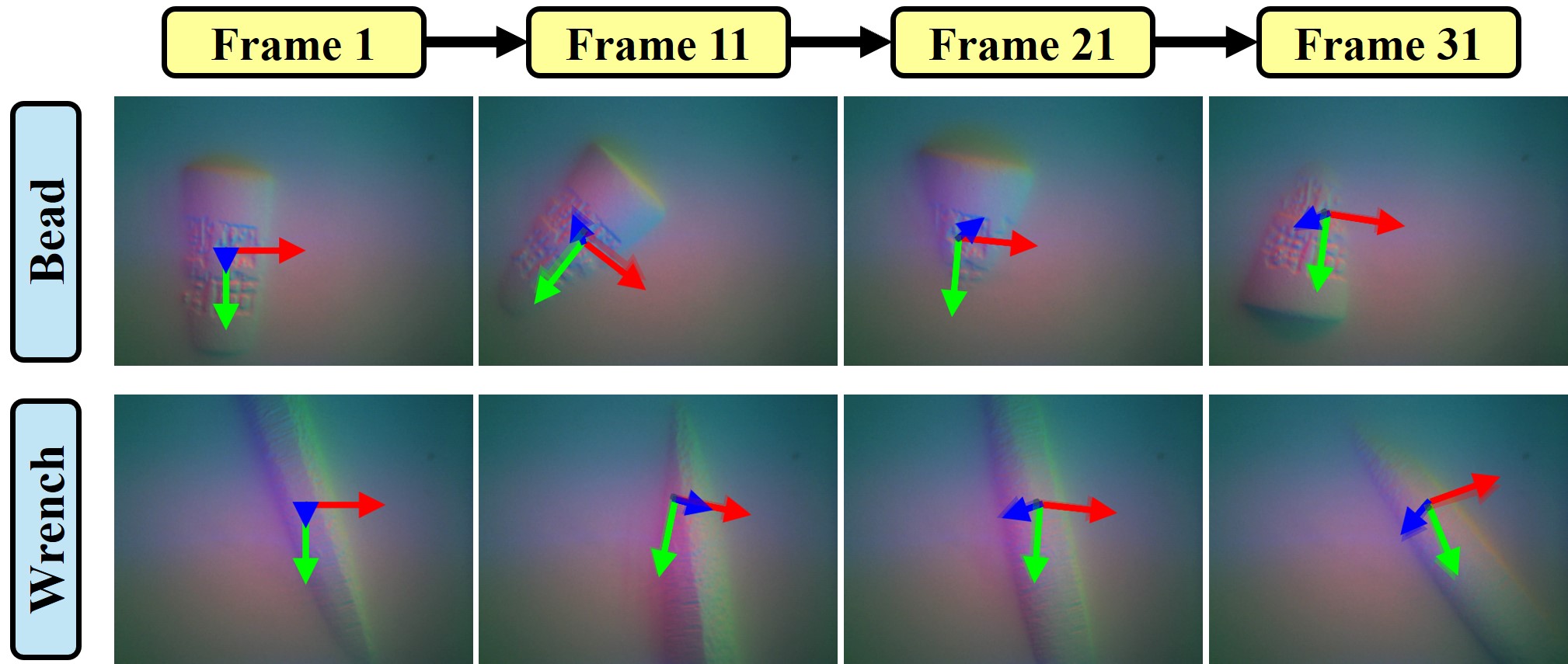}
\caption{Example  trials on two objects. RGB axes show NormalFlow estimated poses. Transparent RGB axes show MoCap poses, nearly overlapping with NormalFlow poses.}
\label{fig:example_trials}
\end{figure}

\subsubsection{Tracking Results}\label{section:tracking_result}
For all methods, we estimate and evaluate the 6DoF sensor pose mean absolute error (MAE) for each frame relative to the first frame. The tracking result is shown in Fig. \ref{fig:tracking_result}, with rotation angles in the z-x-y Euler format.  Two example tracking trials using NormalFlow are shown in Fig. \ref{fig:example_trials}. NormalFlow outperforms baselines on all objects, particularly objects with less textures. We observe that all methods struggled with perfectly symmetrical objects (Ball and Cylinder) due to ambiguities in matching normal maps or point clouds. Excluding these two objects, the average tracking error is shown in Table \ref{tab:tracking_result}. Note that the translational tracking error of MoCap (ground truth) is about $0.2$mm. We find that FPFH+RI often falls into local minima, explaining the challenges of extracting features from tactile point clouds. Meanwhile, ICP consistently performs worse than NormalFlow and FilterReg. While NormalFlow only slightly exceeds FilterReg's performance on highly textured objects (like Avocado and Seed), it significantly outperforms FilterReg on less textured objects (like Wrench and Hammer) by maintaining robust tracking where FilterReg often fails. To the extreme, NormalFlow can robustly track objects like Table, which is considered textureless by human standards.

{\color{revised} \noindent \textbf{Ablation Study}:
To demonstrate NormalFlow's resilience to height estimation errors, we conduct an ablation study by rerunning NormalFlow without height information (assuming a zero height map). The results show only minor changes in 6DoF tracking performance (Table \ref{tab:tracking_result}), with MAE values of  $0.19$mm, $0.18$mm, and $0.18$mm for x, y, and z translations, and $1.2^\circ$, $1.57^\circ$, and $0.69^\circ$ for $\theta_x$, $\theta_y$, and $\theta_z$ rotations.}

%We also show that height map errors affect NormalFlow minimally. 

%This experiment highlights the robustness and accuracy of NormalFlow over the point cloud registration approaches.

\begin{table}[ht]
\centering
\setlength{\tabcolsep}{0.18cm}
\begin{tabular}{|c|c|c|c|c|c|c|}
\hline 
Method & x(mm) & y(mm) & z(mm) & $\theta_x(^{\circ})$ & $\theta_y(^{\circ})$ & $\theta_z(^{\circ})$ \\ 
\hline 
\hline 
NormalFlow & \textbf{0.17} & \textbf{0.18} & \textbf{0.15} & \textbf{1.13} & \textbf{1.42} & \textbf{0.64} \\ 
\hline 
FilterReg & 0.85 & 1.05 & 0.20 & 1.96 & 2.59 & 15.4 \\ 
\hline 
ICP & 1.22 & 3.44 & 0.85 & 2.27 & 3.30 & 15.9 \\ 
\hline 
FPFH+RI & 2.38 & 1.69 & 1.26 & 2.93 & 36.8 & 27.8 \\ 
\hline
\end{tabular}
\caption{6DoF tracking MAE for all objects, excluding the two perfectly symmetrical objects (Ball and Cylinder).}
\label{tab:tracking_result}
\end{table}

\begin{figure}[t]
\centering
\includegraphics[width=0.98\linewidth]{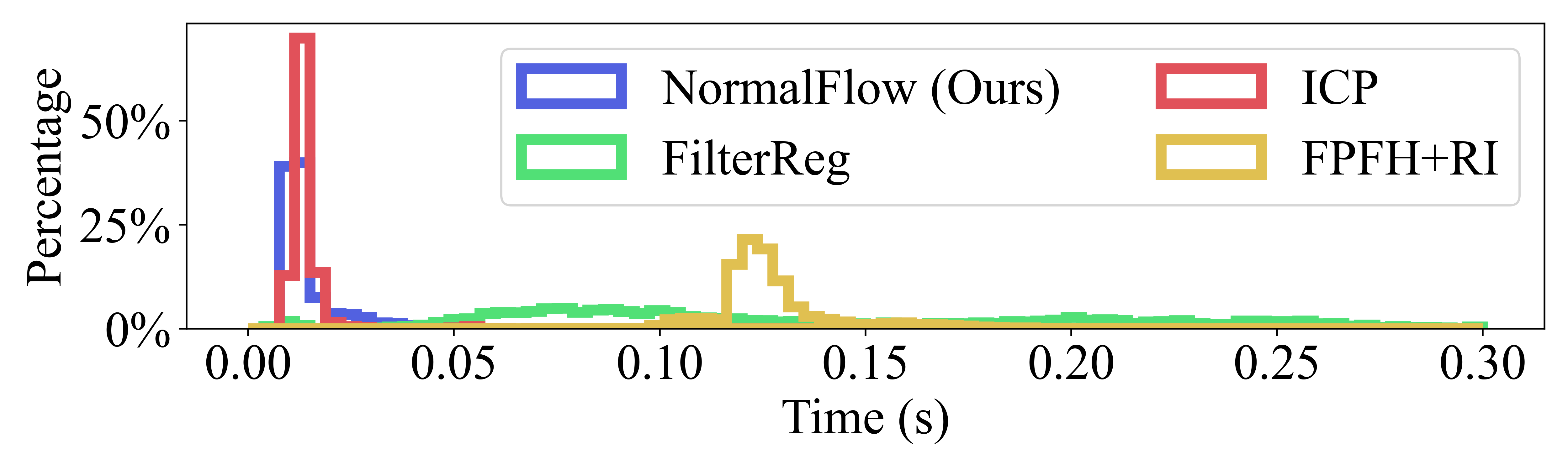}
\caption{The runtime histogram. The average runtimes are: NormalFlow ($13.9$ ms), ICP ($13.6$ ms), FilterReg ($145$ ms), and FPFH+RI ($127$ ms).}
\label{fig:runtime}
\end{figure}

\subsubsection{Runtime Analysis}

In all experiments, we randomly subsample $N =
5000$ pixels when running NormalFlow (Section \ref{section:speed_up}). For a fair runtime comparison in this subsection, we similarly subsample $N=5000$ points in the point cloud for the baseline algorithms. Note that, unlike here, the baselines in all other subsections run using the full point cloud. All algorithms are tested on a laptop equipped with an AMD Ryzen 7 PRO 7840U CPU without GPU acceleration. The runtime histograms are shown in Fig. \ref{fig:runtime}. The average runtime of NormalFlow is $13.9$ ms, closely comparable to ICP at $13.6$ ms, and significantly faster than FilterReg at $145$ ms and FPFH+RI at $127$ ms.

\subsubsection{Long-horizon Tracking Results} 
\begin{figure}[t]
\centering
\includegraphics[width=0.98\linewidth]{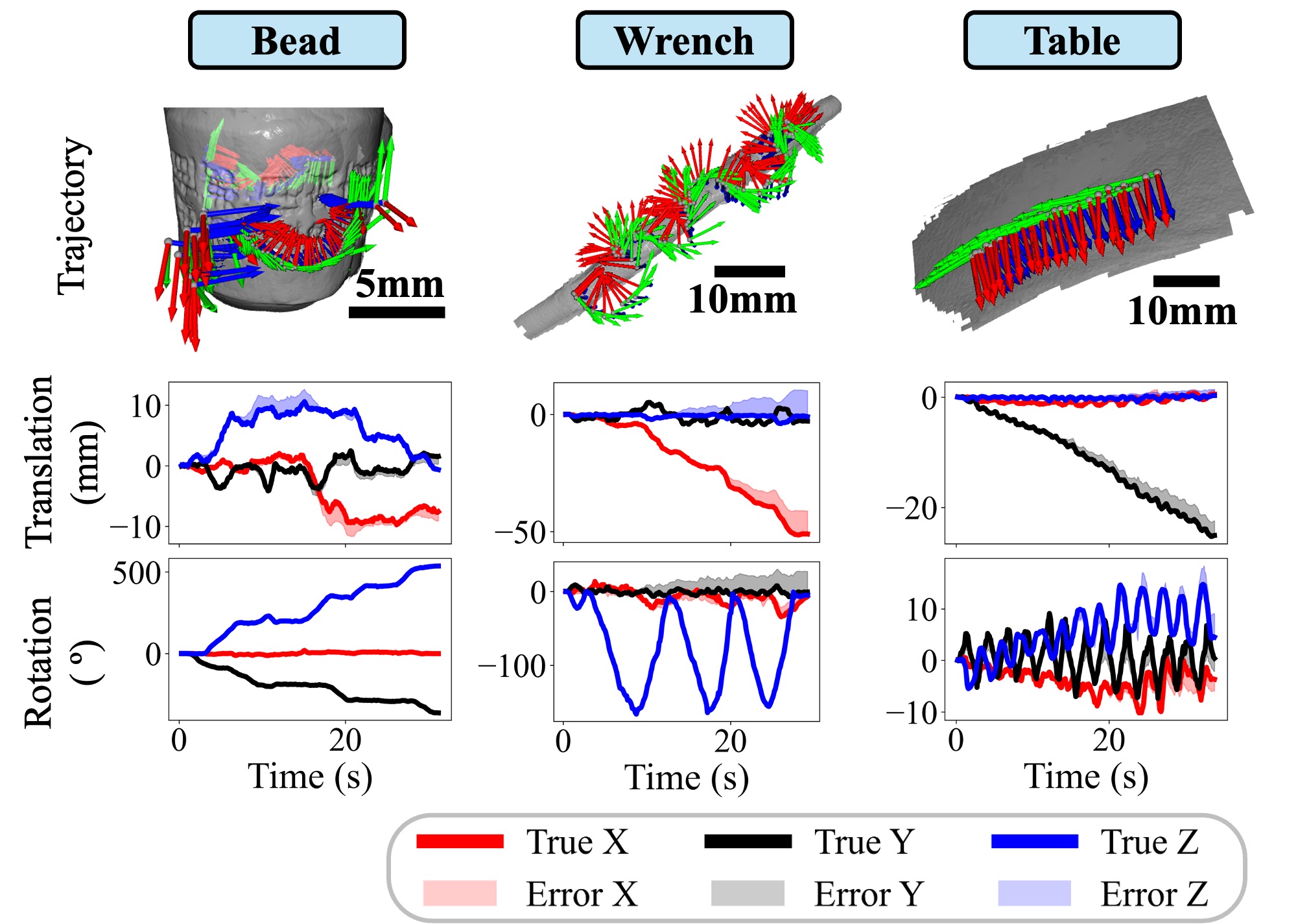}
\caption{Long-horizon tracking results. For each column: \textbf{[top]} sensor trajectories with poses as RGB coordinate axes; \textbf{[bottom]} true 6DoF sensor movements tracked by MoCap (solid lines) and NormalFlow estimation error (shaded area), which is often too small to be seen.}
\label{fig:long_horizon_tracking}
\end{figure}

We demonstrate the long-horizon tracking performance of our approach using three objects with varying tracking difficulty: Bead (easy), Wrench (medium), and Table (hard). For each object, we conduct a trial where the sensor travels a significant distance from its initial pose, requiring multiple keyframes during the tracking process. The NormalFlow tracking results (Fig. \ref{fig:long_horizon_tracking}) indicate minimal tracking error even after extensive movement. After rolling the Bead $360$ degrees along the y-axis and twisting it $540$ degrees along the z-axis, the tracking errors in the y-axis and z-axis remain $2.5$ and $1.8$ degrees, respectively.  Note that determining rolling angles along the y-axis has been previously considered challenging \cite{zhao2023}. {\color{revised}While minor drift occurs, future work can correct it with a factor graph layer \cite{sodhi2022}, with NormalFlow providing more accurate between-factor estimates than conventional ICP methods.}

%The NormalFlow tracking results for the three objects are shown in Fig. \ref{fig:long_horizon_tracking}, illustrating that the tracking error remains nearly zero even after extensive movement. After rolling the Bead in the y-axis of $360$ and twisting in the z-axis with $540$ degrees, the y-axis and z-axis tracking error remains $2.5$ and $1.8$ degrees. Note that determining rolling angles (y-axis) are previously considered very challenging \cite{zhao2023}.

\subsubsection{Tracking by Sliding} 

\begin{figure}[t]
\centering
\includegraphics[width=0.98\linewidth]{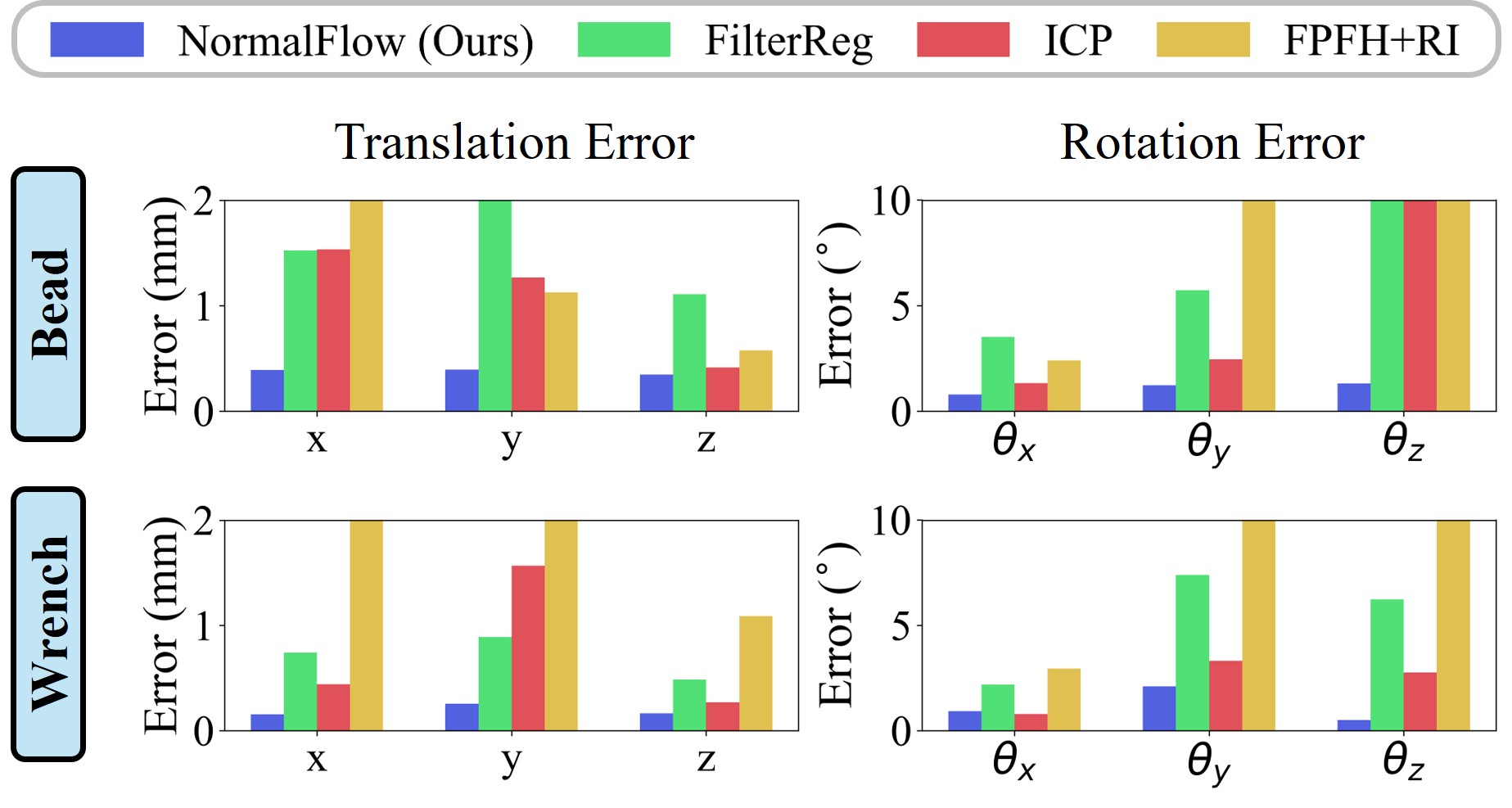}
\caption{6DoF tracking MAE under pure sliding.}
\label{fig:sliding_result}
\end{figure}

Although pure sliding objects on the sensor are discouraged as they wear the Gel quickly, we conducted this experiment to show that NormalFlow can track well under pure sliding. We test the tracking performance under pure sliding with two easily slidable objects: Bead and Wrench. Seven trials were conducted for each object, with an average sliding distance of $13.2$ mm for the Bead and $13.1$ mm for the Wrench, reaching the distance limit without setting new keyframes. The 6DoF tracking MAE of NormalFlow and baselines are shown in Fig. \ref{fig:sliding_result}. The results show that NormalFlow tracks effectively under pure sliding and outperforms the baseline methods.

\begin{comment}
To evaluate the tracking accuracy of NormalFlow, we collect tracking data on $12$ objects (Fig. X). The objects can be classified into three groups, where six objects are daily objects, three objects are small textured objects, and three objects are simple geometrical objects. Four of the six daily objects are from the YCB dataset. During data collection, we clamp the object on a workbench and mount GelSight Mini on a movable plate which is tracked by the OptiTrack Motion Capture system to obtain the group truth poses. Since pure sliding action can easily break the Gel of GelSight Mini, we collect tracking data mainly by rotating on the object's surface (including rolling in the x-y direction and twisting in the z-direction). We make sure that translation from the initial pose is achieved during the rotation actions. For each object, we collect $7$ trials of object movement by touching at a randomly selected place (for hammer and wrench, we sample only from the handle region). In average, each trial includes $80$ images. We make to move not too far from the initial pose so that enough intersection between the contact region is presented so that no keyframes need to be set.

1. Explain the dataset
2. Explain how data is collected and why rolling (and twisting) are preferred. Explain the range of movements. (tools touch on only handles)
3. Explain how ground truth is collected
4. Introduce the baselines
5. Show the result
\end{comment}

\begin{comment}
1. Explain the hardware
2. Explain that we subsample points for baselines
3. Show the result
\end{comment}

\subsection{Movement Tolerance for Tracking} \label{section:convergence}
\begin{figure}[t]
\centering
\includegraphics[width=0.98\linewidth]{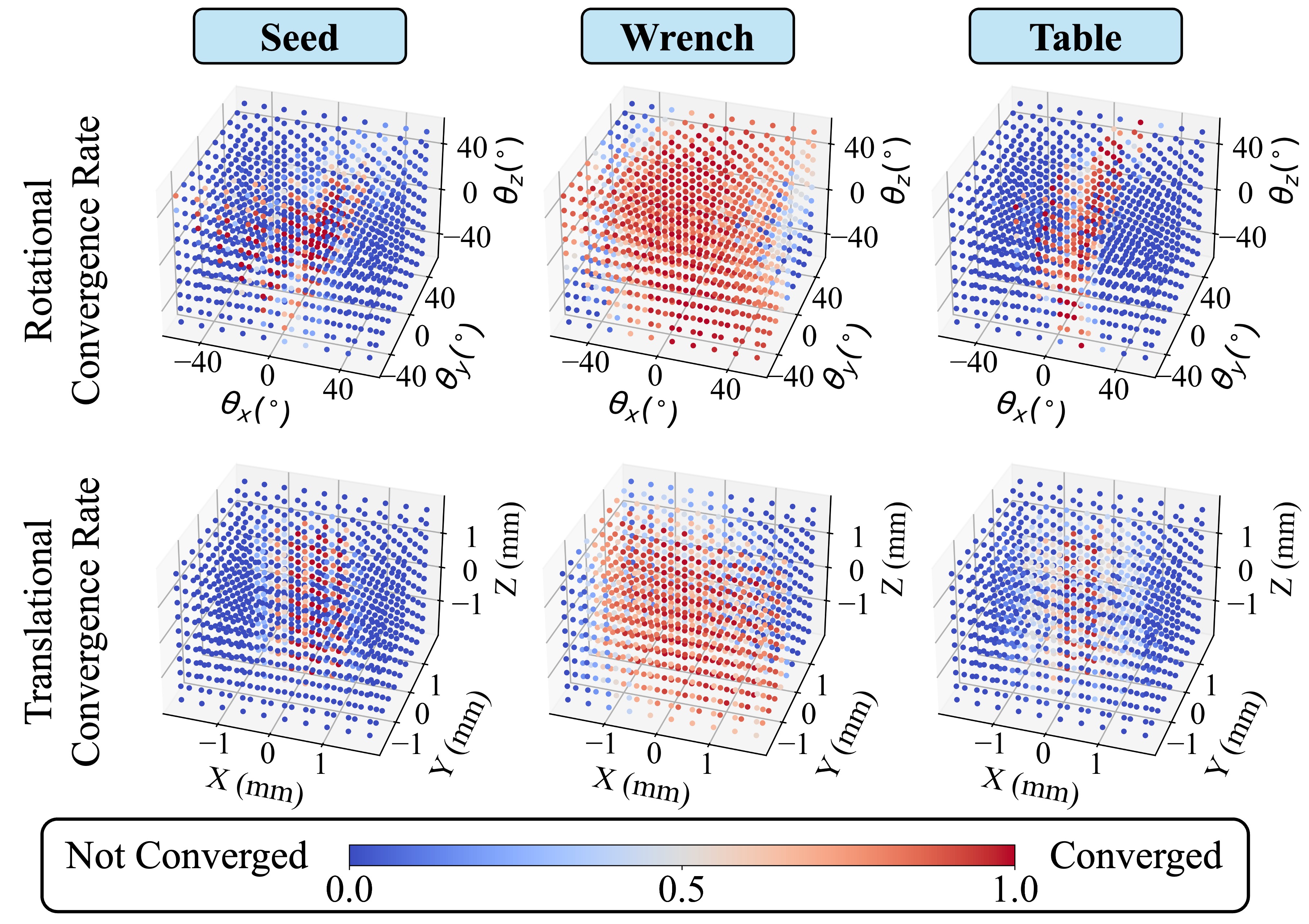}
\caption{Convergence rate for different initialization conditions for the three objects. NormalFlow tracks well when object movement between frames is in the red region.}
\label{fig:convergence_zone}
\end{figure}

When the object moves too quickly, NormalFlow may initialize far from the true pose, causing tracking failures. We evaluate the maximum object moving speed NormalFlow can track by analyzing its region of convergence---the range of pose initialization from which tracking succeeds---using three objects with varying tracking difficulty: Seed (easy), Wrench (medium), and Table (hard). The Wrench is touched and tracked at the neck. {\color{revised} For each object, we conduct a trial, with each tactile image testing $2000$ initial conditions based on the true pose with added shifts.} Of these shifts, $1000$ involves rotations evenly sampled within ±60° across all axes, and $1000$ involves translations evenly sampled within ±2 mm across all axes. Fig. \ref{fig:convergence_zone} shows the convergence rate for different initializations, with Table having the smallest region of convergence. We speculate that objects with less symmetrical global geometry, like Wrench, have a larger region of convergence. On average, NormalFlow tracks Table effectively as long as translational movements between frames stay below $1.39$ mm ($34.8$ mm/s) and rotational movements below $40.1^\circ$ ($17.5$ rad/s). This result shows that NormalFlow is robust to fast rotation but less so to fast translation, which is usually not an issue since fast translation wears the Gel quickly and should be avoided.

\subsection{Resolution Comparison}
\begin{figure}[t]
\centering
\includegraphics[width=0.98\linewidth]{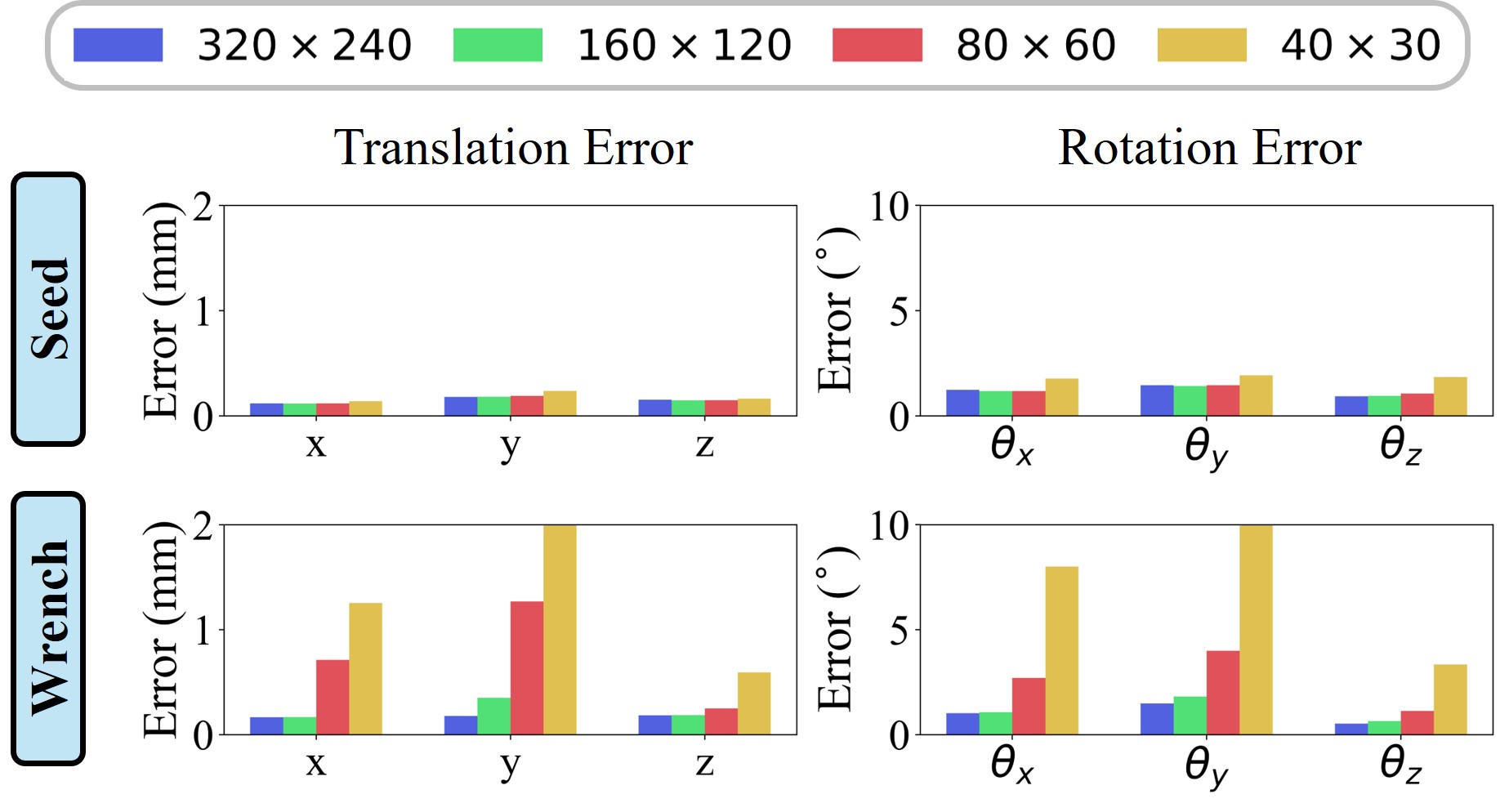}
\caption{Effect of spatial resolution on tracking accuracy.}
\label{fig:resolution_comparison}
\end{figure}

We analyze how sensor spatial resolution affects tracking accuracy. Fig. \ref{fig:resolution_comparison} shows the tracking MAE for Seed and Wrench when using downsampled tactile images. Accurate tracking is maintained with 2x downsampling, but lower resolution degrades performance for Wrench (less textured).

% \textbf{[left]} A sample tactile image at the smallest trackable resolution; \textbf{[right]} 6DoF tracking MAE.

\subsection{Tracking Performance on DIGIT}
\begin{figure}[t]
\centering
\includegraphics[width=0.98\linewidth]{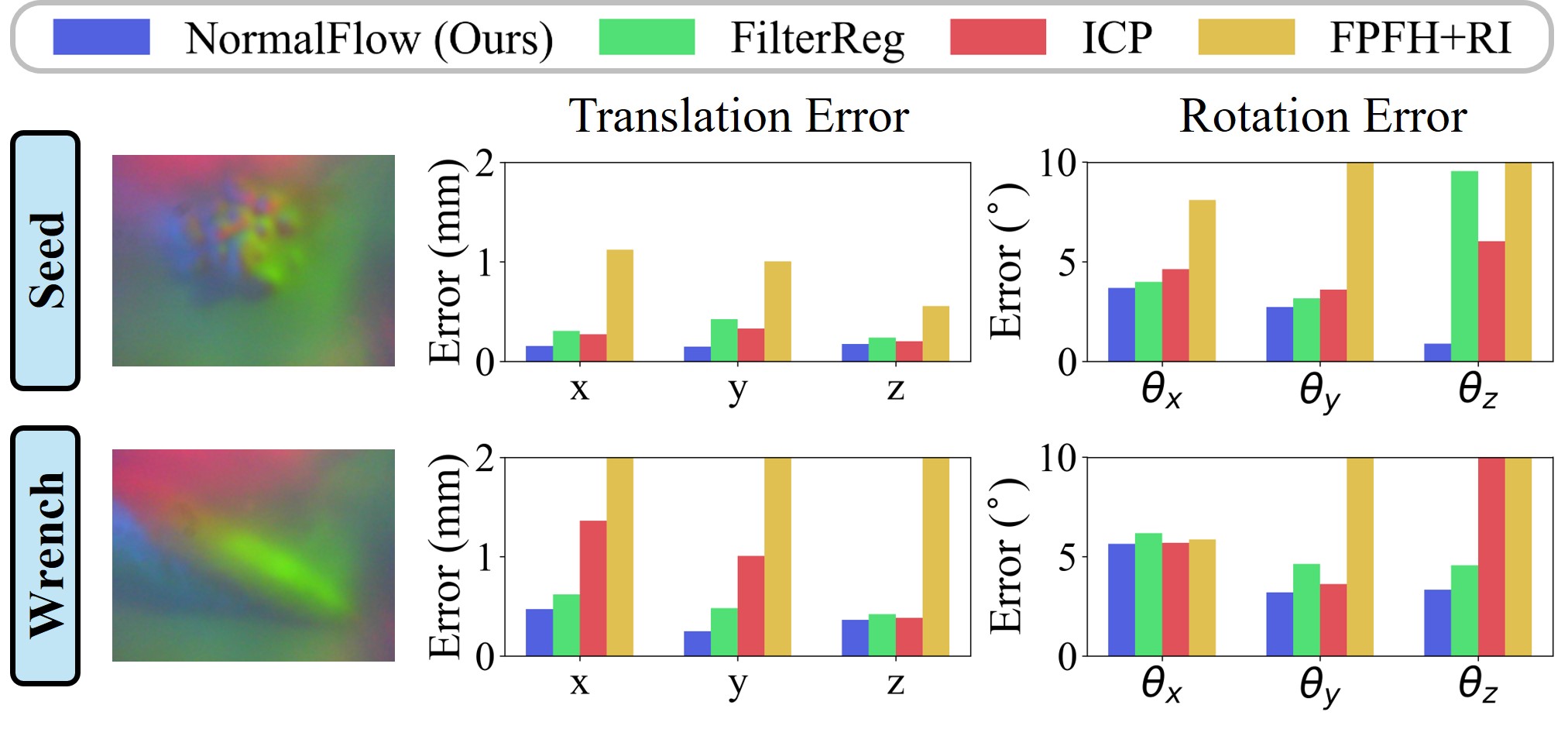}
\caption{DIGIT tracking results for Seed and Wrench. \textbf{[left]} Sample tactile images; \textbf{[right]} 6DoF tracking MAE.}
\label{fig:digit_result}
\end{figure}

We demonstrate our approach using the DIGIT sensor \cite{lambeta2020}. GelSight Mini and DIGIT are the two most common off-the-shelf vision-based tactile sensors. DIGIT has a resolution of $640 \times 480$, a $13 \text{mm} \times 9.7 \text{mm}$ sensing area, and operates at $60$ Hz. Compared to GelSight Mini, DIGIT's blurrier images with less texture detail result in less accurate shape and normal estimation. Using the setup described in Section \ref{section:data_collection}, we collect tracking data on Seed and Wrench with 7 trials each. Fig. \ref{fig:digit_result} shows the 6DoF tracking MAE using all methods. NormalFlow outperforms all baselines, though less so on Wrench due to DIGIT's texture loss.

\section{Application: Tactile-based 3D Reconstruction} \label{section:application}
We demonstrate the power of NormalFlow by applying it to the task of tactile-based 3D reconstruction. Here, we reconstruct the Bead, which poses challenges for vision-based methods due to its carved texture being mistaken for painted texture. We manually roll the Bead on GelSight Mini without breaking contact to collect a tactile video, with each frame revealing a small portion of the Bead (Fig. \ref{fig:example_trials}). Using NormalFlow, we track the 6DoF sensor pose of each frame and set new keyframes as needed. Upon completing a full rotation and returning near the initial pose, we perform loop closure by identifying the frame closest to the initial pose. NormalFlow is then used to determine the relative pose between this frame and the initial one. Finally, we estimate the poses of all frames by optimizing the pose graph with this single loop, setting covariances identically across all factors.

\begin{figure}[t]
\centering
\includegraphics[width=0.98\linewidth]{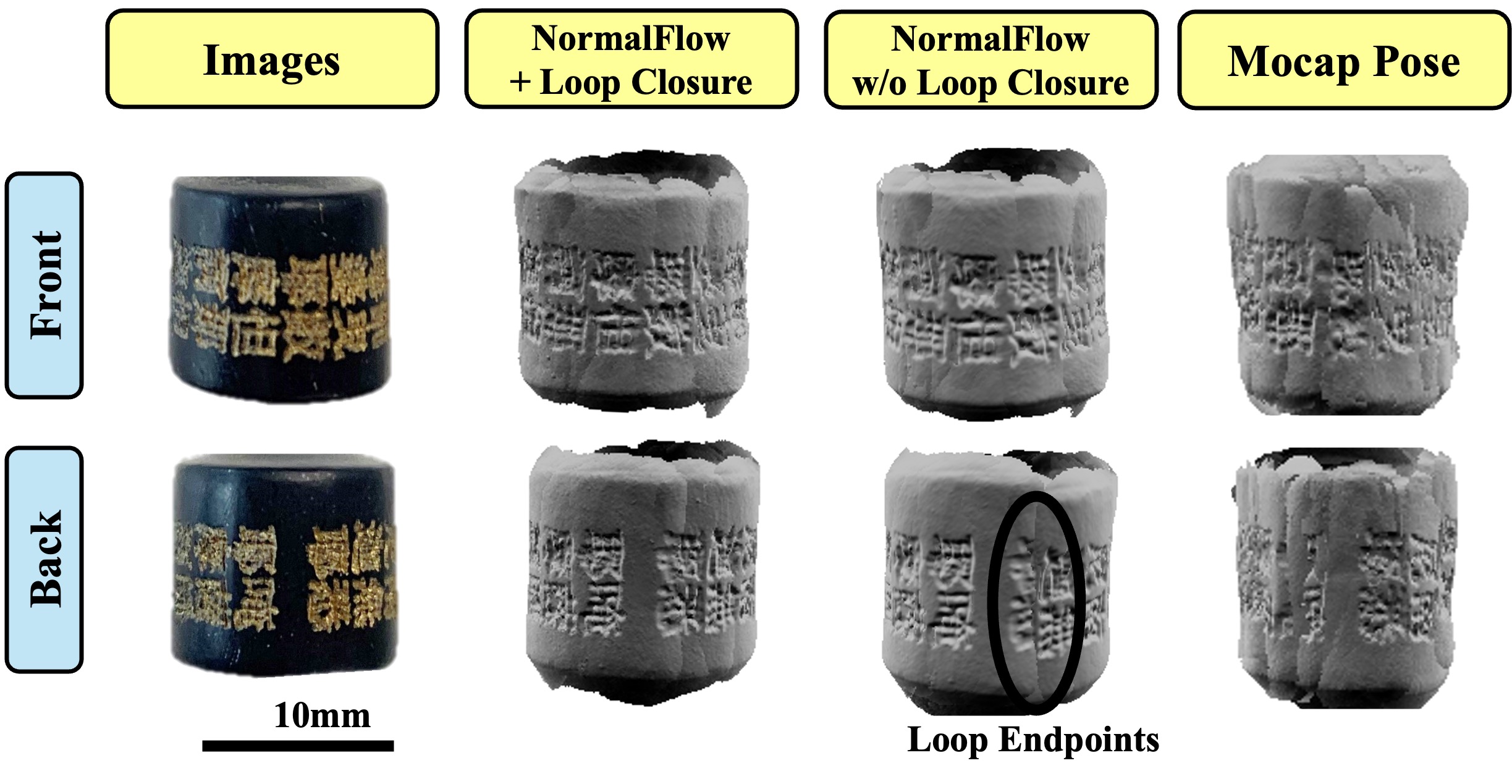}
\caption{
Comparison of tactile-based 3D reconstruction of the Bead using NormalFlow estimated poses (with and without loop closure) versus MoCap tracked poses. {\color{revised} The Bead is rolled on GelSight (trajectory shown in Fig. \ref{fig:long_horizon_tracking}) to capture a tactile video for reconstruction. See the supplementary video for the dynamic process.}
}

%The sensor trajectory is shown in Fig. \ref{fig:long_horizon_tracking}. {\color{revised} See supplementary video for the dynamic process.}}

\label{fig:reconstruction}
\end{figure}

{\color{revised}Fig. \ref{fig:reconstruction} compares three reconstructions of the Bead using local meshes from individual frames registered to NormalFlow poses (both with and without loop closure) and MoCap-tracked poses. The reconstruction result using NormalFlow estimated poses qualitatively outperforms the one using MoCap.} This experiment highlights the accuracy of NormalFlow, demonstrating that even with naive pose graph optimization, NormalFlow achieves significantly better 3D reconstruction results compared to previous methods like \cite{zhao2023}.

\begin{comment}
\subsection{Blindfolded Stick Balancing}

In this task, our goal is to balance a stick on the fingertip of a robot arm using only tactile sensing. We mount a GelSight Mini on a 6DoF robot arm (UR5e by Universal Robotics) to balance a $1$ meter stick, as shown in Fig. \ref{fig:balancing}. Accurate real-time estimation of the stick's orientation is achieved by NormalFlow. We use an LQG control policy to balance. Details are omitted as they are not the main focus. Fig. \ref{fig:balancing} shows a trial of balancing the stick under disturbance, and the balancing video is available in the supplementary material.

\begin{figure}[t]
\centering
\includegraphics[width=0.98\linewidth]{Figures/balancing.jpg}
\caption{Real-time balancing the Wrench on the fingertip using only tactile sensing.}
\label{fig:balancing}
\end{figure}
\end{comment}
\section{Conclusion}
In this work, we present NormalFlow, a fast, robust, and accurate 6DoF pose tracking algorithm for vision-based tactile sensors. By directly minimizing discrepancies between surface normal maps instead of point clouds, our approach outperforms baseline methods and tracks well even with low-texture objects. We also evaluate the effects of object movement speed, sensor choice, and sensor resolution on tracking performance. Finally, we demonstrate the effectiveness of NormalFlow in tactile-based 3D reconstruction. However, NormalFlow is not prone to limitations. NormalFlow can lose track during fast sliding movements (Section \ref{section:convergence}). Simple fixes include initializing NormalFlow with the translation component of the transformation using ICP results or the shift of the contact region's center. We believe NormalFlow can be widely applied, enabling new advancements for high-precision perception, control, and manipulation tasks.

%In this work, we present NormalFlow, a fast, robust, and accurate 6DoF pose tracking algorithm for vision-based tactile sensors. Our key insight is to find the transformation by directly minimizing discrepancies between surface normal maps rather than point clouds. Our approach outperforms baseline methods and tracks well even with low-texture objects. We also evaluate the effects of object movement speed, sensor choice, and sensor resolution on tracking performance. Finally, we demonstrate the effectiveness of NormalFlow in tactile-based 3D reconstruction, achieving state-of-the-art results. However, NormalFlow is not prone to limitations. A limitation of NormalFlow is its susceptibility to losing track during fast sliding movements (Section \ref{section:convergence}). Simple fixes include initializing NormalFlow with the translation component of the transformation using ICP results or the shift of the contact mask center. We believe NormalFlow can be widely applied, enabling new advancements for high-precision perception, control, and manipulation tasks.

% Acknowledgement after accepted
%\section*{ACKNOWLEDGMENT}

\bibliographystyle{IEEEtran}
\bibliography{reference}

\begin{appendices}
\begin{comment}
\section{Jacobian Matrices in Closed-form} \label{appendix:jacobian}
In this section, we present the closed-form Jacobians for the warp function $\mathbf{W}(x,y;\bm{\theta})$ and the rotated normals $\mathbf{R}_{\bm{\theta}}\mathbf{I}(x,y)$ relative to the transformation parameter $\bm{\theta}$:
\begin{align}
\begin{split}
\frac{\partial \mathbf{W}}{\partial \bm{\theta}} =
\mathbf{P}\left[\mathbf{I}_{3\times 3} \Bigg|
\frac{\partial \mathbf{R}_{\bm{\theta}}}{\partial \theta_x}\mathbf{q} \Bigg|
\frac{\partial \mathbf{R}_{\bm{\theta}}}{\partial \theta_y}\mathbf{q} \Bigg|
\frac{\partial \mathbf{R}_{\bm{\theta}}}{\partial \theta_z}\mathbf{q}
\right] \\
\frac{\partial(\mathbf{R}_{\bm{\theta}}\mathbf{I})}{\partial \bm{\theta}} =
\left[\mathbf{0}_{3\times 3} \Bigg|
\frac{\partial \mathbf{R}_{\bm{\theta}}}{\partial \theta_x}\mathbf{I} \Bigg|
\frac{\partial \mathbf{R}_{\bm{\theta}}}{\partial \theta_y}\mathbf{I} \Bigg|
\frac{\partial \mathbf{R}_{\bm{\theta}}}{\partial \theta_z}\mathbf{I}
\right]
\end{split}
\end{align}
Here, both Jacobian matrices lack components for the z-translation dimension, which explains why the z-translation in $\bm{\theta}$ can not be determined using the Gauss-Newton optimization.
\end{comment}

\section{Inverse Compositional Formulation} \label{appendix:inverse_compositional}
%Instead of directly linearizing Equation (\ref{loss}), consider computing $\bm{\mathit{\Delta}} \bm{\theta}$ that minimizes the discrepancies of the transformed reference normal map using the current estimate $\bm{\theta}$ and the transformed target normal map using the small transformation $\bm{\mathit{\Delta}} \bm{\theta}$: 
Instead of directly linearizing Equation (\ref{eq:loss}), we apply the inverse compositional formulation \cite{Baker2004} by considering a small transformation $\bm{\mathit{\Delta}} \bm{\theta}$. We optimize $\bm{\mathit{\Delta}} \bm{\theta}$ by minimizing discrepancies between two transformed normal maps: the reference map transformed by the current estimate $\bm{\theta}$, and the target map transformed by $\bm{\mathit{\Delta}} \bm{\theta}$. The objective is:

\begin{equation} \label{eq:inv_loss}
\sum_{(u,v)\in \overline{C}} \; \Big[\mathbf{R}_{\bm{\theta}}^{-1}\mathbf{I}'(\mathbf{W}(u,v;\bm{\theta})) - \mathbf{R}_{\bm{\mathit{\Delta}} \bm{\theta}}^{-1}\mathbf{I}(\mathbf{W}(u,v;\bm{\mathit{\Delta}} \bm{\theta}))\Big]^2
\end{equation}
where $\mathbf{R}_{\bm{\mathit{\Delta}} \bm{\theta}}$ is the small rotation matrix parameterized with $\bm{\mathit{\Delta}} \bm{\theta}$. The first-order Taylor expansion on Equation (\ref{eq:inv_loss}) yields:
\begin{equation} \label{eq:linearized_inv_loss}
\sum_{(u,v)\in \overline{C}} \; \Big[\Big(\mathbf{R}_{\bm{\theta}}^{-1}\mathbf{I}'(\mathbf{W}(u,v;\bm{\theta})) - \mathbf{I}(u,v)\Big) - \mathbf{J}\bm{\mathit{\Delta}} \bm{\theta}\Big]^2 
\end{equation}
Here, $\mathbf{J}$ represents the first-order terms of Taylor expansion and can be presented in closed-form:

\begin{equation*}
\begin{gathered}
\begin{aligned}
\mathbf{J} ={}
&\nabla \mathbf{I}(u,v) \cdot \mathbf{P} \cdot 
\left[\mathbf{I}_{3\times 3} \Big| \mathbf{J}_x\mathbf{q} \Big| \mathbf{J}_y\mathbf{q} \Big| \mathbf{J}_z\mathbf{q}\right] - \\
&\left[\mathbf{0}_{3\times 3} \Big| \mathbf{J}_x\mathbf{I} \Big| \mathbf{J}_y\mathbf{I} \Big| \mathbf{J}_z\mathbf{I}\right] \qquad \text{where}
\end{aligned}\\[1em]
\setlength{\arraycolsep}{2pt}
\begin{aligned}
\mathbf{J}_x &= \begin{bmatrix}0 & 0 & 0 \\ 0 & 0 & -1 \\ 0 & 1 & 0\end{bmatrix} & 
\mathbf{J}_y &= \begin{bmatrix}0 & 0 & 1 \\ 0 & 0 & 0 \\ -1 & 0 & 0\end{bmatrix} &
\mathbf{J}_z &= \begin{bmatrix}0 & -1 & 0 \\ 1 & 0 & 0 \\ 0 & 0 & 0\end{bmatrix}
\end{aligned}
\end{gathered}
\end{equation*}
%The key advantage of this inverse compositional formulation \cite{baker2001} is that $\mathbf{J}$ is not a function of the current estimate $\bm{\theta}$, which means it can be pre-computed and re-used for every iteration. In fact, the Hessian matrix $\mathbf{H}$ is the runtime bottleneck of the Gauss-Newton optimization and is computed purely based on $\mathbf{J}$, so it can also be pre-computed and re-used.
Independent to the current estimate $\bm{\theta}$, the matrix $\mathbf{J}$ and the computationally intensive Hessian matrix $\mathbf{H}$ can now be pre-computed and re-used in every iteration of the Gauss-Newton optimization, significantly reducing the algorithm's runtime. Finally, we employ the inverse compositional rule to update the current estimate: $\bm{\theta} \gets \bm{\theta} \circ \bm{\mathit{\Delta}} \bm{\theta}^{-1}$ \cite{Baker2004}.

%Independent to the current estimate $\bm{\theta}$, matrix $\mathbf{J}$ can now be pre-computed and re-used in every iteration for the Gauss-Newton optimization. Moreover, the computationally intensive Hessian matrix $\mathbf{H}$ can also be pre-computed from $\mathbf{J}$, significantly reducing the algorithm runtime. Finally, we employ the inverse compositional rule to update the current estimate: $\bm{\theta} \gets \bm{\theta} \circ \bm{\mathit{\Delta}} \bm{\theta}^{-1}$ \cite{Baker2004}.

\begin{comment}
In contrast to the parameters update rule described by Equation (\ref{eq:forward_update}), we employ the inverse compositional rule to update the current estimate:
\begin{equation}
\bm{\theta} \gets \bm{\theta} \circ \bm{\mathit{\Delta}} \bm{\theta}^{-1}
\end{equation}
This update is implemented by representing transformations in homogeneous matrix form and performing the corresponding composition: $\mathbf{T}(\bm{\theta}) \gets \mathbf{T}(\bm{\theta})\mathbf{T}(\bm{\mathit{\Delta}} \bm{\theta})^{-1}$.
\end{comment}

\end{appendices}
\end{document}